\documentclass{article}

\usepackage{arxiv}

\usepackage[utf8]{inputenc} 
\usepackage[T1]{fontenc}    
\usepackage{hyperref}       
\usepackage{url}            
\usepackage{booktabs}       
\usepackage{amsfonts}       
\usepackage{nicefrac}       
\usepackage{microtype}      
\usepackage{lipsum}
\usepackage{graphicx}
\usepackage{mathtools}
\usepackage{booktabs}
\usepackage[dvipsnames]{xcolor}
\usepackage{comment}
\usepackage{caption}
\usepackage{subcaption}
\usepackage{algorithm,algorithmic}

\title{A Simple yet Brisk and Efficient Active Learning Platform for Text Classification}

\author{ Teja Kanchinadam, Qian You, Keith Westpfahl, James Kim, Siva Gunda, Sebastian Seith, Glenn Fung\\
American Family Insurance, Machine Learning Research Group\\
\{tkanchin, qyou, kwestpa, jjkim, sgunda, sseith, gfung\}@amfam.com \\
}

\begin{document}
\maketitle

\begin{abstract}
In this work, we propose the use of a fully managed machine learning service, which utilizes active learning to directly build models from unstructured data. With this tool, business users can quickly and easily build machine learning models and then directly deploy them into a production ready hosted environment without much involvement from data scientists. Our approach leverages state-of-the-art text representation like OpenAI's GPT2 and a fast implementation of the active learning workflow that relies on a simple construction of incremental learning using linear models, thus providing a brisk and efficient labeling experience for the users. Experiments on both publicly available and real-life insurance datasets empirically show why our choices of simple and fast classification algorithms are ideal for the task at hand. 
\end{abstract}
\section{Introduction}
Topic classification and identification have remained a fundamental problem in industries especially when dealing with large corpora of unstructured text, which paved way for an increasing demand for labeled data. Manual labeling such data, for example, is not only a labor intensive and time demanding task but also is subject to regulations and restrictions as only certain business owners have access rights to view such data. In this study, we propose the use of a fully managed active learning service platform deployed in an enterprise level environment which allows business teams to independently curate machine learning models for topic classification.

As stated in \cite{settles2009active}, ``The key idea behind active learning is that a machine learning algorithm can achieve greater accuracy with fewer training labels if it is allowed to choose the data from which it learns.'' Pool-based
active learning uncertainty sampling is the most classic query strategy for choosing the next unlabeled sample to be labeled by an oracle(e.g. a human annotator). A chosen base learner is then updated by the available labeled samples and continues the uncertainty sampling process. Although recent active learning research has adapted to the latest advancement in deep learning, uncertain sampling remains a practical solution that can be widely applied real world applications due to its simplicity and proven robustness. Moreover, most of existing active learning research, both in natural language processing and computer vision tasks,  focused on simulations using well studied open source data sets, while few research presented how real world applications can be benefited from active learning and what is the best practice in deploying active learning processes and models. In fact, 
several difficulties prevent theoretical promising active learning techniques which can be adopted by practitioners, including implementing fast and cost-effective cross-validation algorithms that can be used to tune the algorithms robustly when limited labeled data is available, and in general, utilizing human annotations more efficiently.  In addition to those difficulties, one must consider the active learning system response time when designing practical active learning algorithms, especially retraining the base learner involves constantly increasing labeled data. In a user interaction study \cite{martin1986system}, it was concluded that 0.1 seconds is the appropriate response time for users to feel like their actions are directly causing something to change in the system; response times longer than 0.1 seconds but less than 1 second will make users feel the short delay but won't distract their focus and chain of thoughts; response times longer than 1 second make users feel they are waiting too long and they will grow impatient and may eventually abandon the labeling process. 

In this paper, we present a robust active learning methodology that addresses some of the above difficulties commonly seen in practice. We describe how we deploy the resulting trained model in a cloud-based environment (AWS) where it can be convoked as an API to incorporate its output into other downstream insurance workflows. Specifically, the contribution of this paper is as follows:


\begin{itemize}
\item We propose a simple, fast, and yet robust adaptive tuning method, that we use in this work to optimize our regularization parameter. This method synergizes very well with the active learning paradigm where speed is paramount while achieving competitive performance.

\item We provide an efficient implementation (Python library) of incremental regularized least squares SVM (PSVM, LS-SVM \cite{psvm,suykens2000sparse}) for incremental active learning with the aforementioned fast adaptive regularization parameter tuning that instantaneous cross-validation. 
\item We also discuss our system architecture and deployment methodology of the proposed active learning platform in the context of two innovative insurance-related business applications. \footnote{Code is available at \url{https://github.com/tkanchin/adaptive_reg_active_learning/}}
\end{itemize}
\label{sec:intro}
\section{Methodology}
\label{sec:methodology}
\subsection{Active Learning}
\label{sec: al}
The typical setup for pool-based active learning for classification is as follows: a pool of unlabeled examples $\mathcal{U}$, a pool $\mathcal{L}$ of labeled example-label pairs $(x, y_x)$, an oracle - usually a human annotator that can supply the label of any $x \in \mathcal{U}$, and a query strategy that selects which example $x^* \in \mathcal{U}$ the oracle should label such that $\mathcal{L^*} = \mathcal{L} \cup \{(x^*, y_{x^*})\}$ yields the maximum information gain versus $\mathcal{L}$.

Uncertainty Sampling is an effective and widely used query strategy. It captures the classifier's ($\theta$) uncertainty about the class of $x$, and can be given as: 
\begin{equation}
    x^*  = argmax_{x \in \mathcal{U}} (\mu(x))
\end{equation}
where $\mu(x)$ can be defined as the Shannon Entropy in a classification setting as
\begin{equation}
\label{eq: entropy}
    \mu(x) = - \sum_{c\in \mathcal{C}} p_c(x)\log p_c(x)
\end{equation}
where $\mathcal(C)$ are our possible classes, and $p_c(x)$ is the
probability that our classifier assigns to $x$ having class $c$.

For the rest of the paper, we will use {\it uncertainty sampling}, which is a simple widely-used proven query strategy and that usually produces competitive learning rates in most real-life applications. In the {\it uncertainty sampling} strategy $\mu(x)$ captures the classifier's uncertainty about the class of $x$ \cite{lewis1994sequential}. The intuition is that not much information is gained if a classifier gets a new label with which it already agreed with high certainty. Note for binary classification ($|\mathcal{C}| = 2$), $\mu(x)$ is maximized when $p_c(x) = .5$ for both classes.  For a linear classifier, this is equivalent to finding unlabeled samples that lie closest to the decision hyper-plane and can be implemented in $\mathcal{O}\left(|\mathcal{U}|\log|\mathcal{U}|\right)$ time.

\subsection{Incremental Learning}
\label{sec: inc_learning}
One of the biggest challenges in any active learning framework is that the model has to be re-trained at every iteration of the process. Typically, a model retrains quickly at the starting iterations as only a few labeled samples are available to train the model. However, as more samples come by, training time usually increases considerably since most learning algorithms complexity depends on the number of training samples. In order to make training time grow slowly across all the iterations, we propose the use of an incremental learning mechanism that takes advantages of the simplicity of a PSVM / LS-SVM classifier to make efficient incremental updates of the model instead of calculating the model form scratch at each iteration.

least-squares-based SVM classifiers are a modified version of standard SVM where a least squares cost function is proposed so as to obtain a linear set of equations in the
either the primal (PSVM) or the dual space (LS-SVM). \cite{suykens2000sparse,psvm}.\\ \\
More succinctly, the goal is to learn a linear classifier $f(x)=\theta'x$, such that $\||f(X)-Y||^2+||\theta||^2$ is minimized. The closed form PSVM solution  with a linear kernel, is similar to a Ridge regression solution and it can be written as:
\begin{equation}
\label{eq:lssvm}
    \theta = (X^TX+\lambda I)^{-1}X^TY
\end{equation}
where $X \in R^{m \times n}$ with $m$ rows (datapoints), $n$ is the number of features and $Y$ is a vector of binary label with values in  \{$+1$, $-1$\}.
$\theta$ are the coefficients of the separating hyperplane; $\lambda$ is the regularization parameter. 
\\
Let $M_n^{-1} = (X^TX+\lambda I)^{-1}$, such that $\theta = M_n^{-1} \sum\limits_{i=0}^{n-1} x_i y_i$.

Note that when a new training point $(x_n,y_n)$ is added, the linear classifier weights can be updated incrementally as follows,
\begin{equation}
\label{eq:inc_final}
    \hat\theta = \theta + M_{n+1}^{-1}x_n(y_n - x_n^T\theta)
\end{equation}

Where the new inverse $M_{n+1}^{-1}$ is a rank-1 update of the previous one and can be calculated very efficiently using the Sherman-Morrison-Woodbury \cite{DENG20111561} formula as follows:
\begin{equation}
    M_{n+1}^{-1} = M_n^{-1} - M_n^{-1}x_n(1 + x_n^TM_n^{-1}x_n)^{-1}x_n^TM_n^{-1}
\end{equation}
where $M_n^{-1}$ is the inverse at previous iteration and $(x_n,y_n)$ is the new point.The inverse can be updated with a low computational cost since the most expensive operation is a matrix-vector multiplication. We will call this incremental process \textbf{LS-SVM SMW}.

Based on some ideas from the batch least squares, equation (\ref{eq:lssvm}) can be written as

\begin{equation}
\label{eq: poly_1}
    \hat\theta = \theta + M_{n+1}^{-1}
    \frac{1}{t}\sum_t x_t(y_t - x_t^T\theta)
\end{equation}
where $t$ are the number of iterations in the incremental learning process. 

If we were to consider not storing and updating the inverse each time, assign $M_{n+1}^{-1} = \gamma_n$, now equation (\ref{eq: poly_1}) can be implemented in $O(1)$ time as it is only requires a vector-vector multiplication. Note that $\gamma_n$ is a scalar and it can chosen by $\gamma_n = n^{-\alpha}$ where $\alpha \in (0.5,1.0)$. This heuristic was proposed in 
\cite{polyak1992acceleration} where the idea is to average trajectories at each step, such that the solution can be obtained by averaging over all previous solutions. We call this process \textbf{LS-SVM Poly}

For both incremental methods described above, during each iteration of the active learning process, the regularization parameter $\lambda$ is fixed and thus cannot be changed when new training points are added for incremental learning. Since the training set is constantly and drastically changing (especially in early iterations when the labeled training dataset is small) the regularization parameter $\lambda$ has to be tuned at every iteration to optimize performance. This process can add additional computing costs that could result in slower iterations of the active learning process.  

In order to address this problem, we consider two approaches. 
\begin{enumerate}
    \item To use Singular Value Decomposition (SVD) to factorize $X$. This factorization of $X$ allows to find close-form solutions for $\theta$ as a function of $\lambda$.
    However, we also have to update incrementally the SVD factorization in each iteration.
    \item We propose the use a simple greedy adaptive tuning regularization approach that uses properties of equation \ref{eq:lssvm} to quickly find values of $\lambda$ that optimize our leave-one-out cross-validation performance over a finite set of candidate $\lambda$'s.
\end{enumerate}

\subsection{Efficiently tuning $\lambda$ using SVD}
\label{sec:svd_lambda}
Let SVD on $X = USV^T$ where $X \in R^{m \times n}$; $U$ and $V$ are orthonormal matrices and $U \in R^{m \times m}$, $V \in R^{n \times n}$; $S$ is a diagonal matrix $S \in R^{m \times n}$;
Equation (\ref{eq:lssvm}) can now be reformulated as follows:
\begin{equation}
\label{eq: svd_lssvm1}
    \theta = (VSU^TUSV^T+\lambda I)^{-1}VSU^TY
\end{equation}
Note that the properties of SVD state that, $U^TU = I$ \& $V^TV=I$. Thus, the above equation can be further reduced to,
\begin{equation}
\label{eq:svd_lssvm2}
    \theta = V(S^2 + \lambda)^{-1}SU^TY
\end{equation}
where $(S^2 + \lambda)$ is a diagonal matrix and its inverse can be cheaply calculated.

When a new point $(x_n,y_n)$ is added, the $U,S,V$ matrices are updated using a SVD rank update discussed in \cite{brand2006fast} and solve for equation (\ref{eq:svd_lssvm2}). Note that even though this method allows us to change $\lambda$ in the incremental learning process, it is still computationally expensive because SVD rank update involves decomposing a re-diagonalized matrix.

\subsection{Solving for $\lambda$: Adaptive regularization Approach}
\label{sec: adap_reg_approach}
We propose a simple yet effective approach to update the regularization parameter $\lambda$ in this approach. The basic idea is to train several parallel incremental learners each initialized with a $\lambda$ parameter. The best learner at each step is identified by performing a leave-one-out cross validation.

\textbf{Generalized Cross Validation (GCV):} The running time for leave-one-out cross validation increases linearly as training set grows in incremental learning. Since, the underlying learner is LS-SVM, \cite{GolubHW07} proposed a simple method to estimate the LOOCV error and is given by:
\begin{equation}
    GCV(\lambda) = \frac{1}{n} \sum_i ((y_i - \hat y_i) / 1 - tr(H)/n)^{2}
\end{equation}
where $\hat y_i$ are the predictions of the model, $y_i$ is the ground truth and $H$ is the hat matrix and is defined as 
\begin{equation}
    H = X(X^TX + \lambda I)^{-1}X^T
\end{equation}

\begin{algorithm}[H]
\begin{algorithmic}[1]
{\small
\caption{Greedy adaptive regularization approach}
\STATE Let $\hat\theta(\lambda^{*}) = \{\theta^{1}(\lambda^{1}), \theta^{2}(\lambda^{2}), ..., \theta^{l}(\lambda^{l})\}$
\STATE Let $\hat g = \{g^{1}, g^{2}, ..., g^{l}\}$ be the corresponding GCV 
\IF {TRAINING}
    \STATE Let $(x_n,y_n)$ be the new training point 
    \STATE Update all learners in $\hat\theta(\lambda^{*})$
    \STATE Update $\hat g$
\ELSE
    \STATE return best model as $\hat\theta [argmin(\hat g)]$
\ENDIF
}
\end{algorithmic} 
\end{algorithm}
\label{alg: alg1}
Note that the set $\{\lambda_1, \dots, \lambda_l\}$ defines the search space, however the optimal value of $lambda$ can change between iterations.

The goal of the greedy adaptive regularization method is to prevent overfitting when tuning the linear models between iterations. Since overfitting commonly occurs on small training sets which is usually the case in active learning scenarios, adaptive regularization can enable the active-learning-produced models to achieve better generalization, especially at early iterations, and hence to produce better queries.
\section{Offline Validation}
\label{sec:experiments}

\subsection{Dataset}
\label{sec:dataset_details}
To simulate the active learning process and to test the design choices, we have used the IMDB movie reviews dataset which consists of 1000 positive and 1000 negative movie reviews. These reviews were labeled and annotated by human annotators and are best described in \cite{zaidan2007using}.

\subsection{Feature Representation}
\label{sec: feat_repr}
To test the performance of our proposed algorithm against various feature representations, we have considered the following:
\begin{enumerate}
    \item \textbf{W2V:} Unweighted  average  of  the  word vectors that comprise the sentence or document \cite{Wieting:2016}. We we have used the publicly available Glove model \cite{pennington2014glove} trained on news articles.
    \item \textbf{USE:} Pre-trained transformer based Universal Sentence Encoder described in \cite{cer2018universal} to map input to vector representation.
    \item \textbf{BERT:} Pre-trained transformer based model described in \cite{devlin2018bert} to map input to vector representation.
    \item \textbf{GPT2:}  Pre-trained transformer based model described in \cite{radford2019language} to map input to vector representation.
\end{enumerate}


\subsection{Incremental Active Learning Results}
\label{sec: inc_al_learning}
We have benchmarked the speed and performance of various formulations against IMDB movie reviews dataset discussed in Section \ref{sec:dataset_details} and the results are shown in Table \ref{inc_speed_benchmarks}. The table compares incremental learning to that of the traditional learning where the learner is re-trained with all the points at each iteration of the learning process. It is evident from the table that the SVD methods are computationally expensive as they involve some heavy operations in the decomposition of matrices. The other LS-SVM methods are computationally cheap, it is important to observe that the LS-SVM Poly is fastest when compared to all the other methods as here we approximating the inverse update operations. However, since the approximation is stochastic in nature, the performance wouldn't be exact when compared to the baseline LS-SVM. Note that the LS-SVM SMW method achieves the best AUC performance while performing around 1000 updates in 1.6 seconds. This roughly translates to negligible computation times when updating the models during the labeling process.

\begin{table}
\centering
{\small
\caption{Table showing the performance comparison between various learning approaches. \textit{LS-SVM SVD} refers to section \ref{sec:svd_lambda}, \textit{LS-SVM SMW} and \textit{LS-SVM Poly} refers to section \ref{sec: inc_learning}}
\begin{tabular}{c c c c} 
\toprule
\textbf{Model} & \textbf{Incremental} &\textbf{AUC} & \textbf{Speed (s)} \\
 \midrule
		LS-SVM & No &0.8588 & 27.32 \\ 
		\textbf{LS-SVM SMW} & \textbf{Yes} & \textbf{0.8588} & \textbf{1.629} \\ 
		LS-SVM Poly & Yes & 0.8289 & 0.072 \\ 
		LS-SVM SVD & Yes & 0.811 & 628.42 \\
\bottomrule
\end{tabular}
\label{inc_speed_benchmarks}
}
\end{table}

\begin{figure*}[!htb]
     \centering
     \begin{subfigure}{.30\textwidth}
         \centering
         \includegraphics[width=\linewidth]{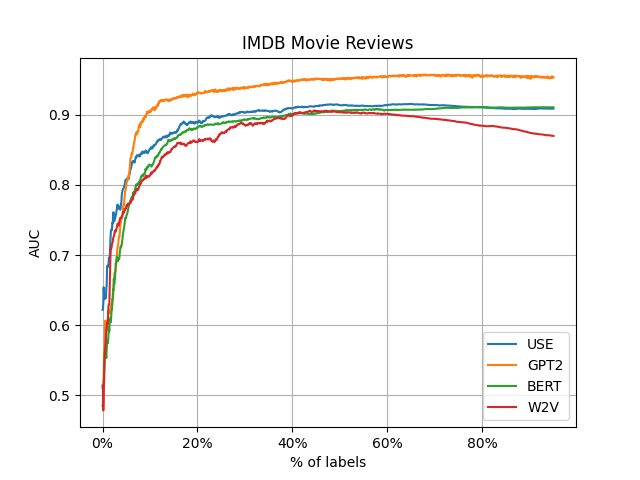}
         \caption{}
     \end{subfigure}
     \begin{subfigure}{0.30\textwidth}
         \centering
         \includegraphics[width=\linewidth]{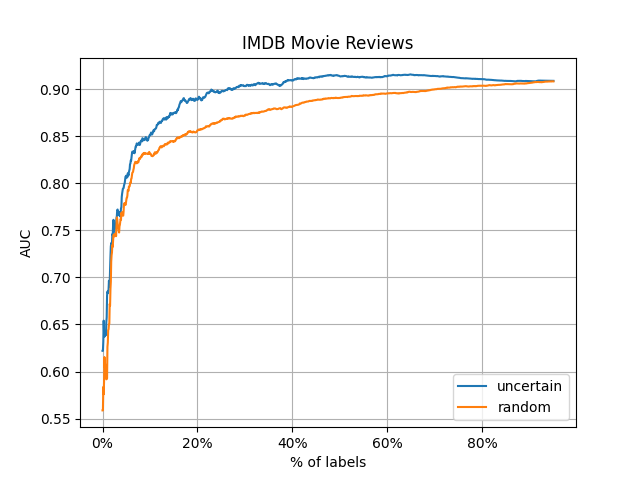}
         \caption{}
     \end{subfigure}
     \begin{subfigure}{0.30\textwidth}
         \centering
         \includegraphics[width=\linewidth]{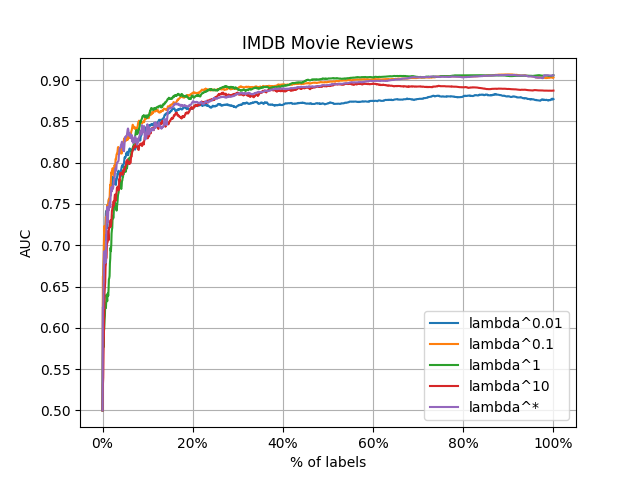}
         \caption{}
     \end{subfigure}
        \caption{Active Learning results on IMBD movie reviews dataset (a) text representation (b) uncertainty sampling versus random sampling (c) performance of adaptive regularization}
        \label{fig:imdb_experiments}
\end{figure*}

\textbf{Active Learning Process}
We have used a incremental LS-SVM classifier as the underlying classification model. We initialize the model with 2 randomly selected labeled examples to start the active learning simulation. In each iteration, we include one example from the unlabeled set to update the linear model until the unlabeled set is completely exhausted. We also calculate the receiver-operating-characteristic (ROC), or an AUC score of the model against a hold out test set. Metrics are reported by averaging the results from 10 runs and for the simplicity of display, we have not reported standard deviations. 
\\
In Figure \ref{fig:imdb_experiments}(a), we show that the proposed incremental LS-SVM model with adaptive regularization and GPT2 features clearly outperform other text representations in an active learning setting using uncertainty sampling.

Similarly, in Figure \ref{fig:imdb_experiments}(b), we show that uncertainty sampling clearly outperforms random sampling using adaptive regularized LS-SVM model.

In Figure \ref{fig:imdb_experiments}(c), we show the performance of adaptive regularization in an active learning setting using uncertainty sampling. It is evident from the figure, the best model at every iteration of the training process is yielding the optimal performance.  


\section{Platform And Applications}
\begin{figure*}[htbp]
	\centering
		\includegraphics[scale=0.23]{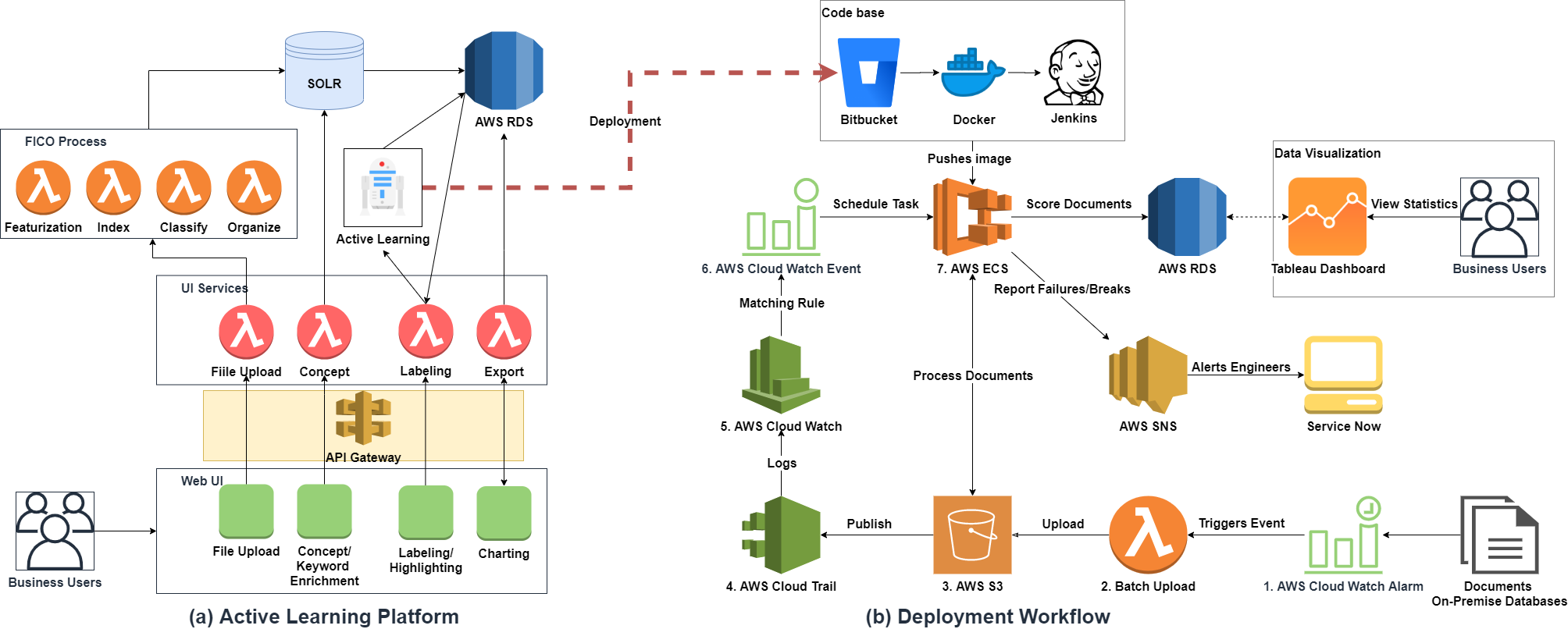}
	\caption{Active Learning Platform and Model Deployment Workflow}
	\label{fig:rocket_deployment}
\end{figure*}
Based on the learning methodology discussed in Section \ref{sec:methodology}, we have developed an active learning platform for users to upload and label unstructured documents (see Figure \ref{fig:rocket_deployment}(a)). Once the documents are labeled and models are curated, they are directly deployed into a production ready hosted environment (see Figure \ref{fig:rocket_deployment}(b)). 
The rest of this section is divided as follows: a) In Section \ref{sec: rocket platform}, we introduce our active learning platform. b) In Section \ref{sec: app_for_insurance}, we discuss about two innovative insurance domain specific applications. c) In Section \ref{sec:deployment}, we briefly discuss our deployment methodology.


\subsection{Active Learning Platform} 
\label{sec: rocket platform}
We have developed an active learning platform with an interactive web-based user interface (UI) and all the services supported by this UI are orchestrated by a back-end server, designed using a light-weight and independent micro-services architecture \cite{thones2015microservices}. 



\begin{figure*}[h]
     \centering
    \begin{subfigure}[b]{0.4\linewidth}
          \includegraphics[width=\textwidth,height=0.5\textwidth]{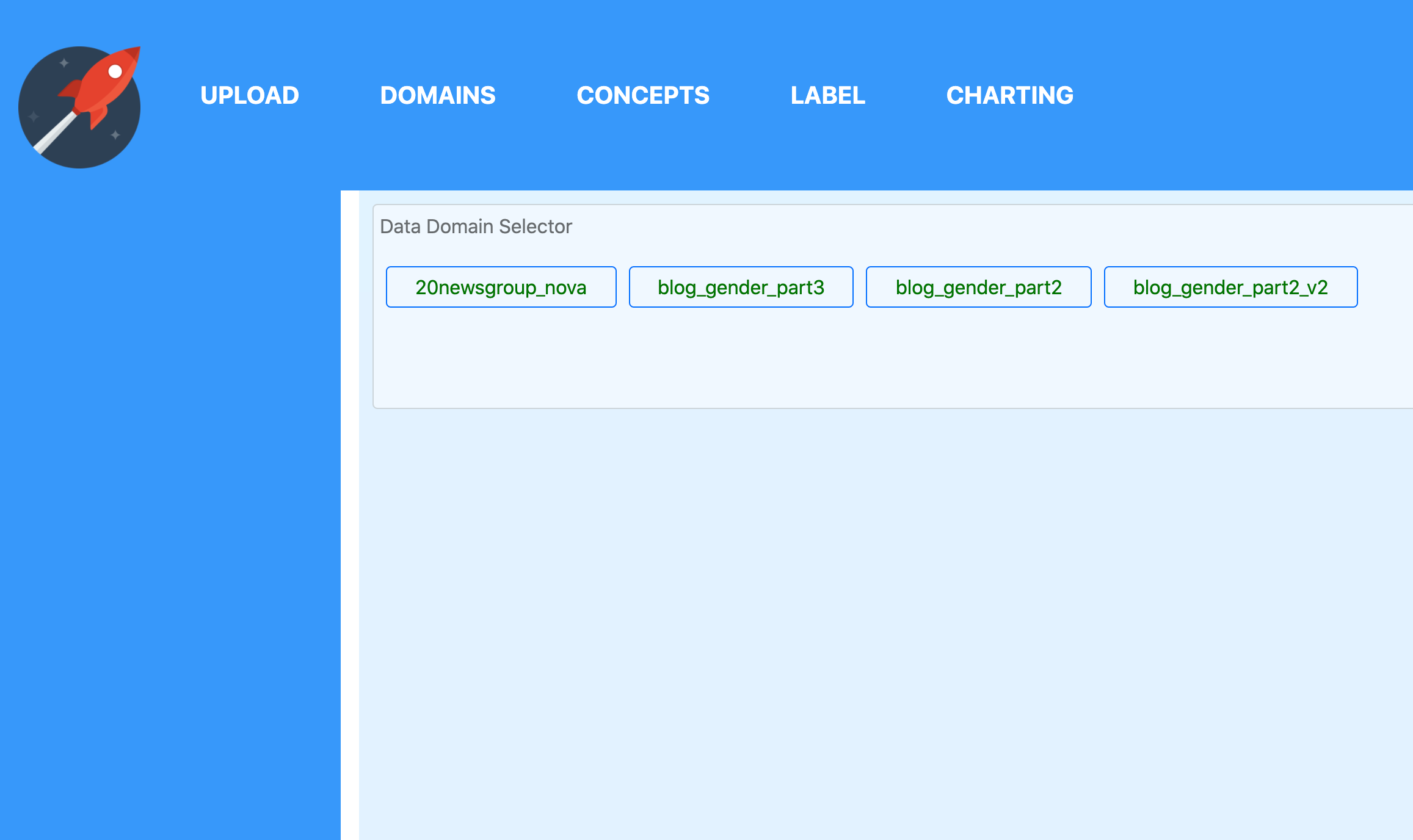}
        \subcaption{UI: File Upload}\label{fig:1a}
        \end{subfigure}
        \begin{subfigure}[b]{0.4\linewidth}
         \includegraphics[width=\textwidth,height=0.5\textwidth]{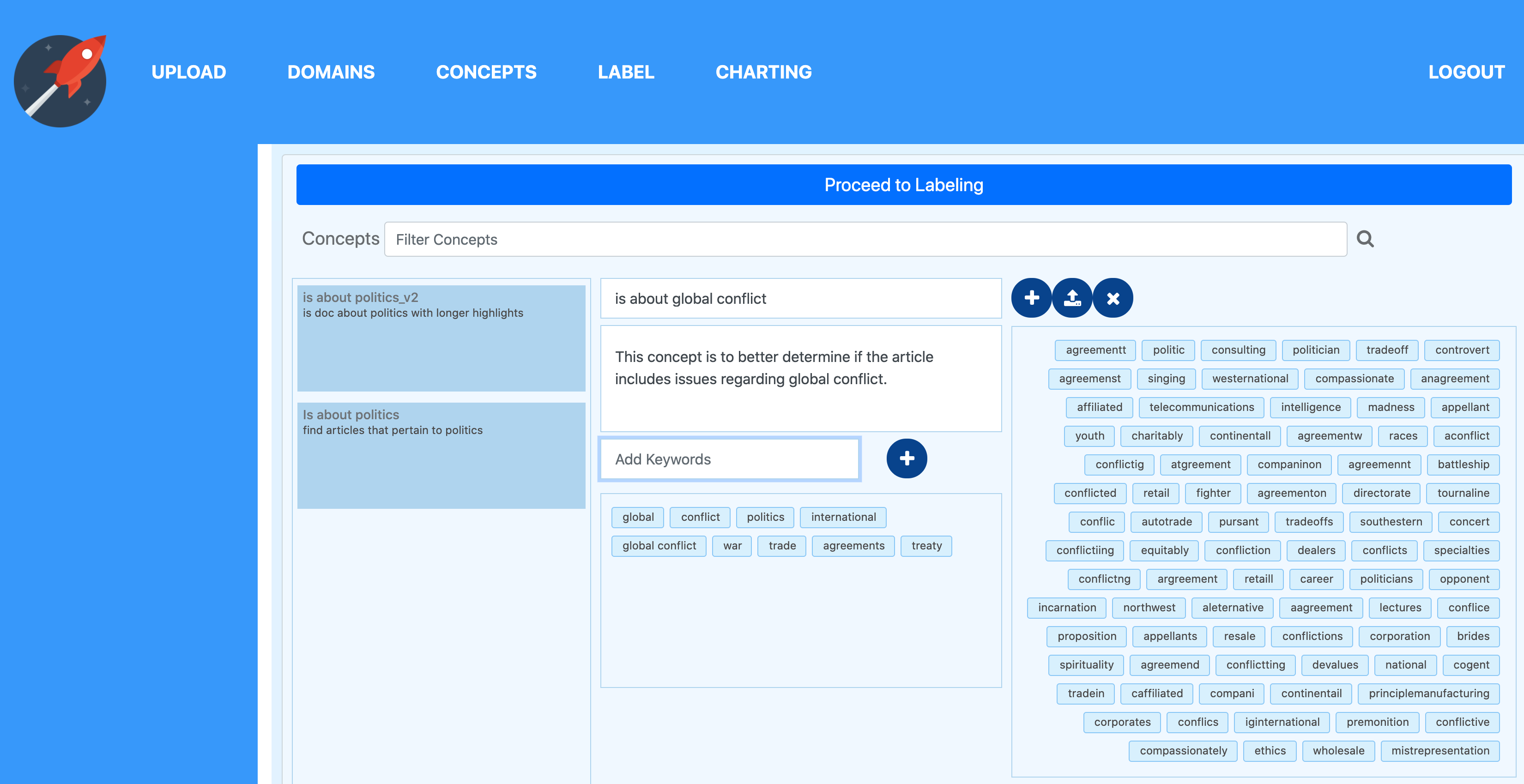}
         \subcaption{UI: Concept Definition}\label{fig:1b}
      \end{subfigure}
      \\
      \begin{subfigure}[b]{0.4\linewidth}
         \includegraphics[width=\textwidth,height=0.5\textwidth]{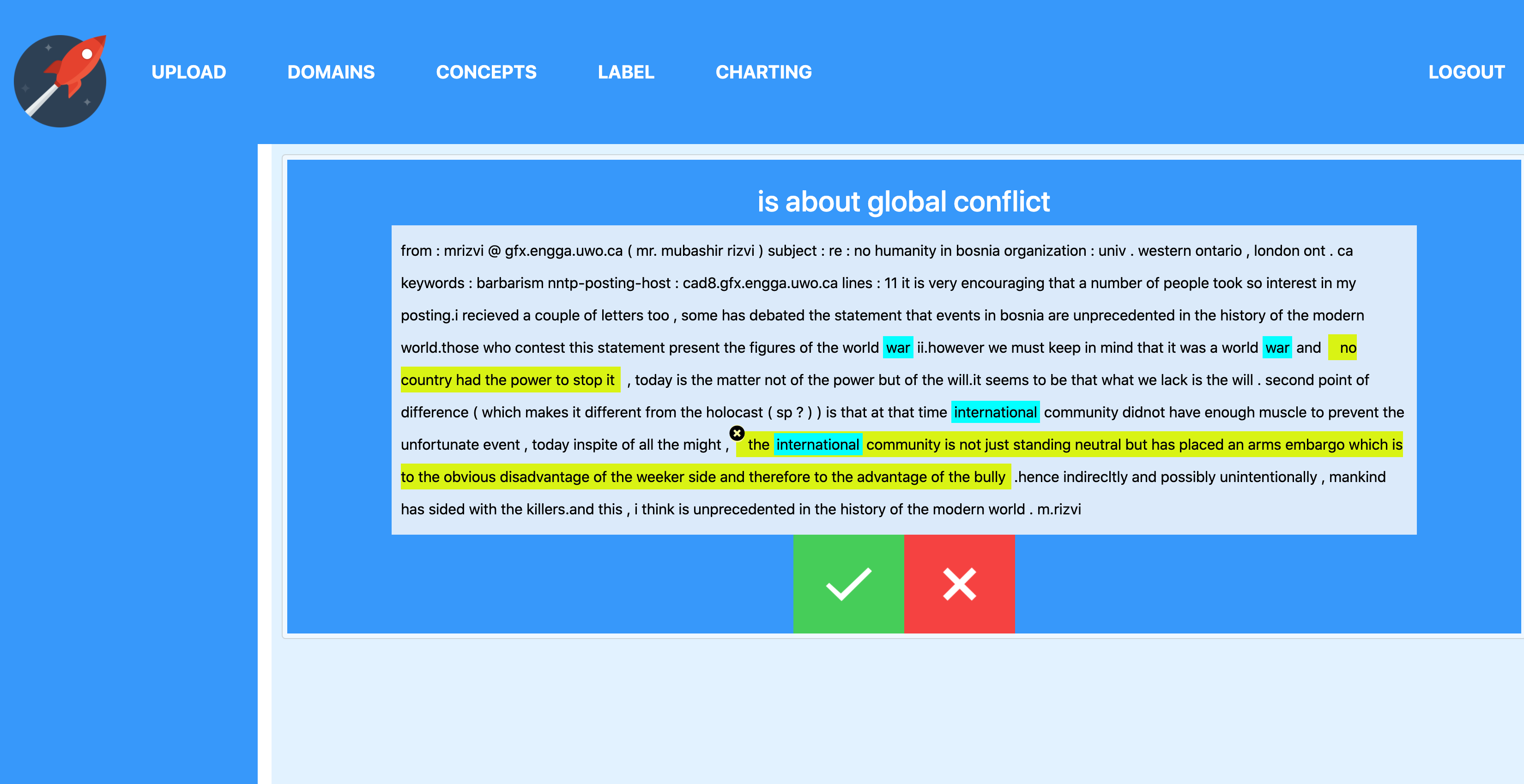}
         \subcaption{UI: Labeling Interface}\label{fig:1c}
     \end{subfigure}
       \begin{subfigure}[b]{0.4\linewidth}
         \includegraphics[width=\textwidth,height=0.5\textwidth]{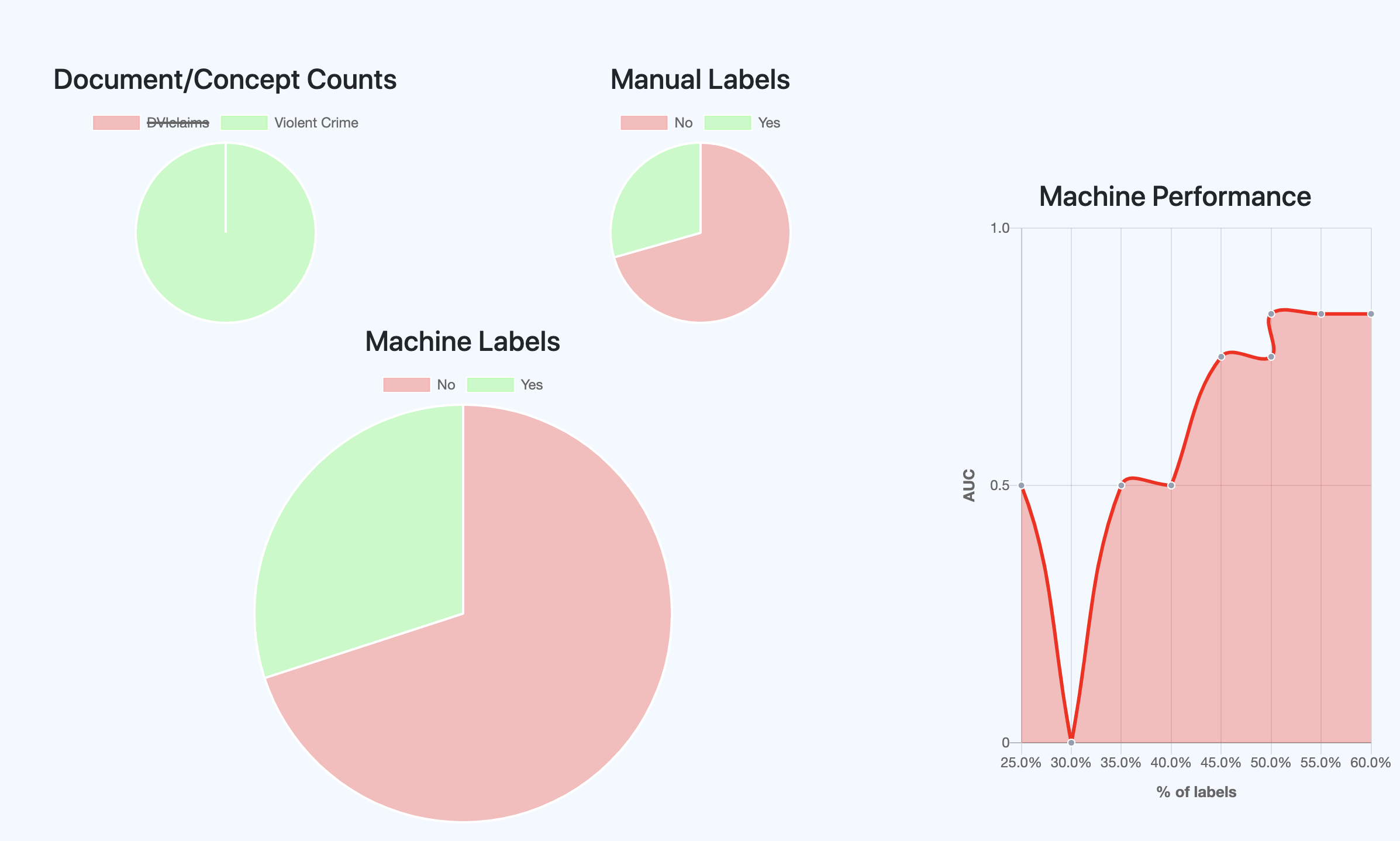}  
         \subcaption{UI: Labeling Statistics}\label{fig:1d}
     \end{subfigure}
    
        \caption{Interactive Web UI interface of the proposed Active Learning Platform }
        \label{fig:rocket_fig}
\end{figure*}

\begin{enumerate}
    \item In Figure \ref{fig:rocket_fig}(a), once the user uploads a dataset of interest to be analyzed, we automatically process this dataset using \textit{FICO} which is a series of steps namely $\{Featurization, Indexing, Cleaning, Organization\}$.  
    \begin{enumerate}
        \item \textit{Organization:} There are several different types of insurance data that can flow into our system. In this step, we identify the source of the data and store them in a database.
        \item \textit{Cleaning:} In order to deal with the potential exposure of personal identifiable information (PII), we use regular expressions to filter out the email addresses, phone numbers, ids, people names and other characters. In addition to this, we use regular expressions to identify the numerical entities such as time, date, dollar signs, etc. and assign special tokens to them.
        \item \textit{Indexing:} In order for us to query our corpus for any documents containing a set of tokens/keywords, we index our data using Elasticsearch \cite{gormley2015elasticsearch}, which provides the flexibility and scalability necessary to perform a token-based search over millions of documents in near real-time.
        \item \textit{Featurization:} In this step, our goal is map our input to a text representation that is well suited for learning algorithms with limited labeled data. We employ a K80 GPU Amazon Web Services (AWS) service to extract features from a GPT2 model discussed in Section \ref{sec: feat_repr}.
    \end{enumerate}
    \item In Figure \ref{fig:rocket_fig}(b), the user defines a concept they wish to extract from the documents of interest using a short list of keywords that are related to that concept. The active learning platforms offers the ability to enrich this set of keywords by supplementing semantically similar words using a word embedding based model. We have trained a FastText model described in \cite{joulin2017bag} using many insurance-related text datasets available within the company to learn a rich word embedding that captures the language commonly used in the insurance domain. 
    \item In Figure \ref{fig:rocket_fig}(c), users can label documents as positive or negative for the defined concept. The UI will also allow users to highlight certain phrases/keywords so as to explain the rationale behind their decision. Note that the documents presented to the user for labeling are provided by an active learning algorithm powered by incremental adaptive regularized method. This makes the model update and document query to negligible computation time providing a smooth labeling experience to the users.
    \item In Figure \ref{fig:rocket_fig}(c), users can track the progress and performance of their labeling tasks. It would enable users to a) track the labeling progress b) model predictions on the remaining documents c) AUC performance at different steps in the labeling process. Note that every $3^{rd}$ labeled document (randomly selected) is used as a hold-out set to calculate AUC.
    \end{enumerate}

\subsection{Applications for Insurance}
\label{sec: app_for_insurance}
\subsubsection{Auto Claims Loss Causes}
Insurance business processes generate large amounts of unstructured text which are packed with relevant information. For example, every time we receive a claim, the request and a brief hand-typed description of the incident is stored in our internal database systems for business related purposes. Here is an example of an auto claim:
\par
\begin{flushleft}
``{\it The front right tire on the trailer blew out and broke apart causing damage to the wheel well and possibly to the wiring on the electronic brakes for that wheel.}''
\end{flushleft}
\par
In order to produce metrics related to the different claim causes, product analysts within our company manually categorize these claims using simple heuristic rules like keywords and regular expressions. This is not only a time consuming and labor intensive job but it is also prone to human subjection and variance.  This problem can be addressed by the proposed active learning platform where a) an analyst can upload documents of interest related to claims b) define a concept for the desired loss cause c) label a few documents to train models to automatically classify incoming claims into the appropriate loss cause d) deploy these models into production and automate the categorization process. 

\noindent Table \ref{rv_label_percentage} shows the results of this application on identifying loss causes i.e. \textit{Tire Blowout}, \textit{Water Damage}, and \textit{Wind Damage}, and curating machine learning models using the proposed active learning platform. From a total of 7287 auto insurance claims, the loss cause \textit{Wind Damage} has 4350 documents retrieved using simple heuristics like keywords; analysts labeled a total of 1804 samples from these retrieved documents indicating whether the claim belongs to loss cause or not. Since every $3^{rd}$ labeled document is used as a hold-out set, the machine learning model curated using active learning achieved an AUC of $94.3$. \textbf{model $+^{ve}$} is obtained by applying the machine learning model to the total number of retrieved candidates. This gives an insight that the the model was able to effectively prune a good amount of the retrieved documents and only retain those that are more likely belonging to the desired loss cause.

\subsubsection{Property Claims Audit}
Property claims are unstructured textual narratives containing information about building characteristics, utilities and surrounding environment. Auditors would like to audit these claims periodically to verify that claims have been operated using standard procedures. Currently, the auditing process is conducted by randomly selecting a subset of claims. This can be addressed more efficiently using our proposed active learning platform. For example, if the auditors want to audit the claims related to \textit{"structural damage"}, they can curate a model using our proposed active learning platform and deploy the model to production. The model is then applied to a huge corpus of claims generating predictions, the auditors can rank the predictions of the model and audit the claims which are more likely related to \textit{"structural damage"}.



Table \ref{div_label_percentage} shows the results of our application on identifying topics for property claims auditing. The topics are developed from a corpus of around 140,000 property claims. From the table, most of these audit concepts are relatively rare. For example, there are only eight \textit{"Pollution"} claims which is a small percentage. Instead of combing through hundreds of thousands of claims to find \textit{"Pollution"} claims, the auditor can identify such claims quickly and efficiently with the proposed framework.


\subsection{Deployment}
\label{sec:deployment}



The architecture of our deployment workflow is shown in Figure \ref{fig:rocket_deployment}(b). Our workflow is a highly scalable event-driven system using AWS managed services operating at minimum costs. Our deployment workflow is explained as below:
\begin{enumerate}
    \item First, we package the models along with the application code in a docker container and push the image to a container registry on the cloud. We then use Elastic Container Service (ECS) which provides a pay-per-use container orchestration engine without the maintenance of computing infrastructure to run the packaged docker image as an ECS task. This setup enables the system to vertically scale by increasing the memory and vCPUs as needed. 
    \item Second, we create batches of documents and store them on S3 using a Lambda service, which provides a pay-per-use serverless computing infrastructure. 
    \item Third, a CloudWatch is triggered as an ECS task once the batched documents are uploaded to S3. The ECS task reads the documents from S3, runs it against the application code and store the results in an AWS Managed Relational Database Service (RDS). 
\end{enumerate}













\begin{figure*}
     \centering
     \begin{subfigure}[b]{0.4\textwidth}
         \centering
         \includegraphics[width=.9\textwidth]{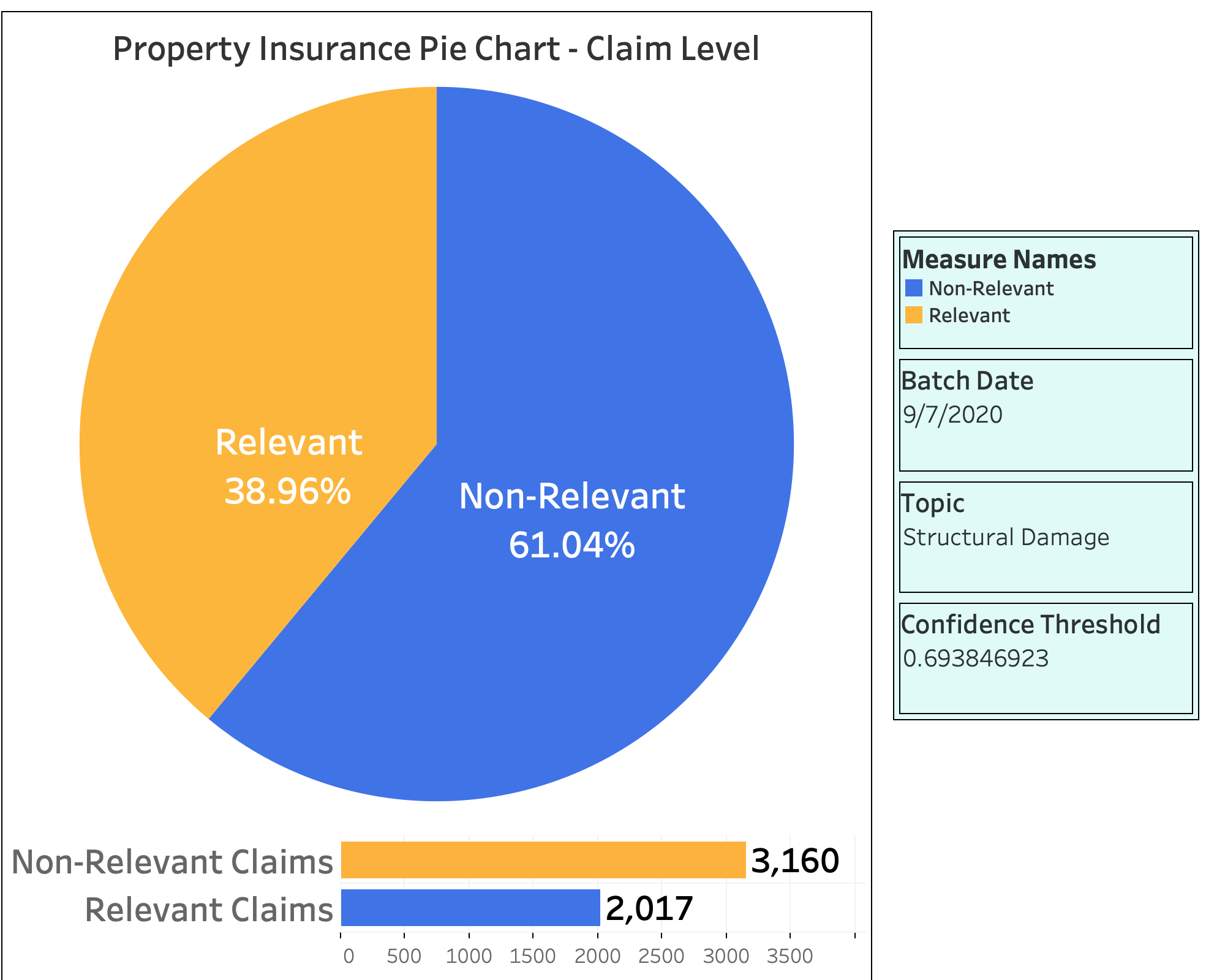}
         \caption{View of claims by topic with respective confidence threshold}
         \label{fig:Tableau_Pie_Chart}
     \end{subfigure}
     \hfill
     \begin{subfigure}[b]{0.4\textwidth}
         \centering
         \includegraphics[width=.9\textwidth]{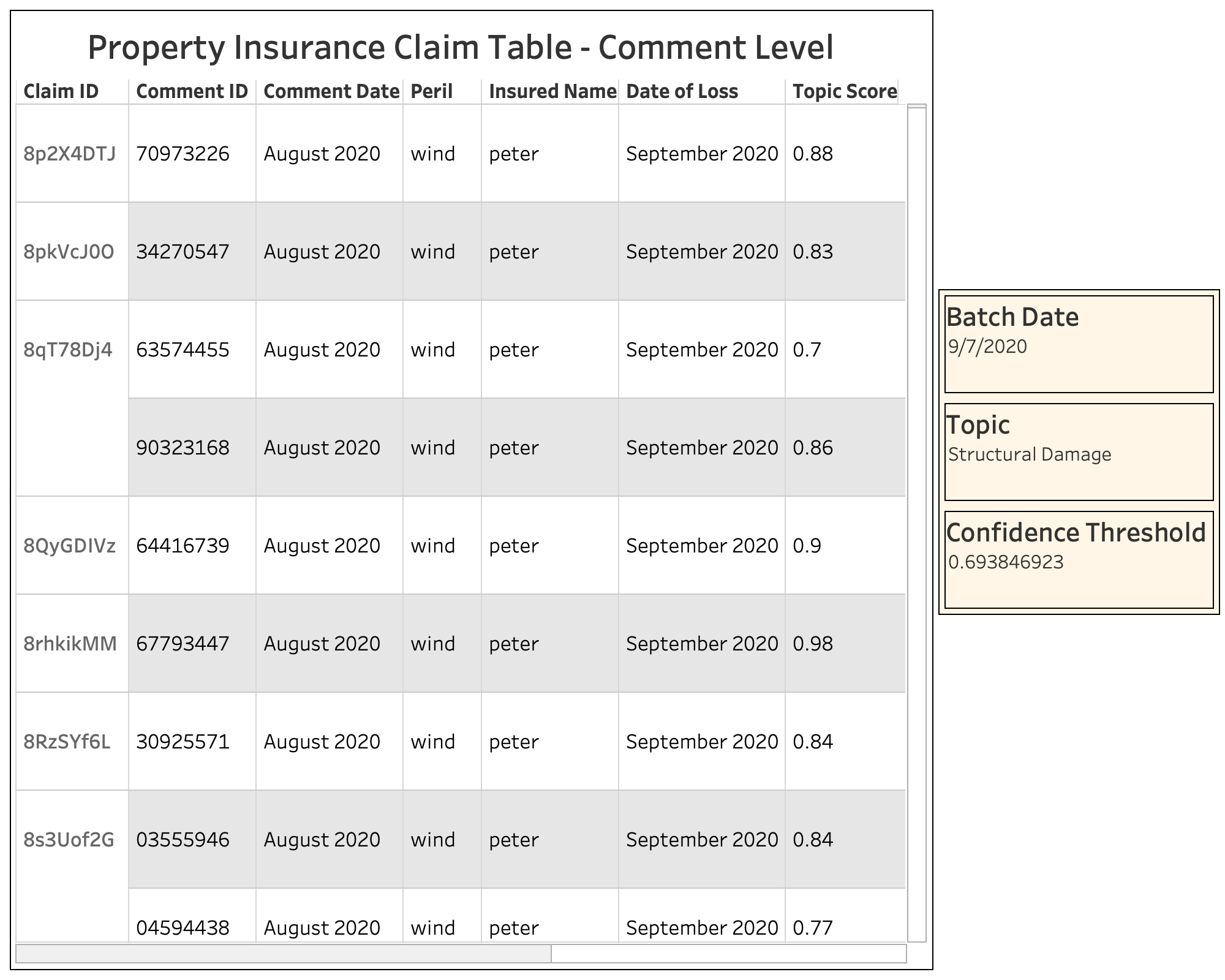}
         \caption{Detailed view of claims by topic with respective confidence threshold}
         \label{fig:Tableau_Table}
     \end{subfigure}
        \caption{Interactive Business Dashboard for Property Claims Audit}
        \label{fig:tableau_fig}
\end{figure*}

Business users can interactively visualize the results via dashboards (see Figure \ref{fig:tableau_fig}) in real-time. For example, in Figure \ref{fig:tableau_fig}(a), the dashboard provides percentage breakdown of non-relevant claims versus relevant claims to be audited. In Figure \ref{fig:tableau_fig}(b), users are able to view the relevant claims at a detailed level as well as export the contents for further investigation. 



\begin{table}
\centering
{
\caption{Results for Auto Claims Loss Cause}
\begin{tabular}{c c c c c} 
\toprule
 RV topics & No. retrieved candidates & No. human labels & AUC & model $+^{ve}$\\
 \midrule
Wind Damage  & 4350 & 1804 & 94.3 & 701\\
Tire Blowout &  1490 & 768 & 89.4 & 264\\ 
Water Damage & 1453 & 520 &  91.3 & 197\\ 
\bottomrule
\end{tabular}
\label{rv_label_percentage}
\title{Example RV claim loss causes, retrieved documents, labels during active learning and results}}
\end{table}


\begin{table}
\centering
{
\caption{Results for Property Claims Audit}
 \begin{tabular}{c c c c c} 
 \toprule
 Property claim topics & No. retrieved candidates & No. human labels & AUC & model $+^{ve}$\\
 \midrule
 Structural Damage &1326 & 231 &  81.54 & 239\\ 
Pollution  &1000 & 200 & 72.85 & 8\\ 
 Asbestos &  341 & 67 & 84.61 & 71\\
 Smoke Type & 224 & 69 & 81.25 & 24\\
 Water Intrusion &  798 & 161 & 72.85 & 82\\
\bottomrule
\end{tabular}
\label{div_label_percentage}
\title{Example Property Insurance Claim Audit Concepts, retrieved document, labels during active learning and results}}
\end{table}

\section{Related Work}
\label{sec:rel_work}
\subsection{Active Learning}
Active learning aims to develop label-efficient algorithms by sampling the most representative queries to be labeled by an oracle, which usually is a human annotator. 
Many sampling strategies have been developed over the past decades \cite{info-active-learning,em-pool-al,qbc-al,svm-active-learning,lrank-al}. 
The most effective and commonly used pool-based active learning is probably uncertainty sampling \cite{al-wse,baseline-al,structured-pred}. Recently developed deep active learning also research on how to adapt new model architectures to uncertainty sampling \cite{lloss-al,bald,dbal-nlp}. 
Although deep models can out-perform classic uncertainty sampling, they are usually computationally inefficient.


\subsection{Incremental Active Learning}

A deployed active learning system needs to retrain and respond in real time for users to keep focused and efficient. Therefore we based our LSSVMS formulation on least square SVM in \cite{psvm} and \cite{SuykensV992,SuykensV99} which produces the solutions by solving a set of linear equations  of solving a more costly constrained quadratic programming. Under the formulation of LSSVMs,  \cite{DENG20111561} proposes a generalized Sherman–Morrison–Woodbury formula which can be represented in Moore–Penrose inverse and the generalized Drazin inverse forms. 
On the other hand, more research \cite{ZHOU2015717,brand2003fast} proposed to incrementally update Singular Value Decomposition (SVD) based on incoming data.
In this paper, we explored using SVD to factorize our least-square solutions to allow for the incremental update of the weights as well to optimally tune the regularization parameter in every step of the updates.
\section{Conclusion and Future Work}
\label{sec:conclusion}
We have designed, implemented and deployed a system that solves a problem relevant to the insurance domain, although the system could be use in in any other domains where concepts need to be asserted or classified from an unstructured text source. Our system optimally incorporates feedback form human labeler following an active learning strategy. The system is designed such that the human feedback is incorporated into model training in an instantaneous way so it seamlessly enhances the user experience. 
The machine learning methodology behind this implementation is simple yet elegant and relies on classical results form the least-squares literature. In particular, numerical empirical evidence has shown us that the greedy adaptive regularization approach is of great value to the system and the reason it is possible to use it is a consequence of relying on simple linear models rather than state-of-the-art more complex architectures where performing cross validation would be prohibitive because of both the time constraints and the small size of the available training data characteristic in active learning scenarios.  Infrastructure for large-scale deployment of our system is also explained in detail and parts of the code will be shared for reproducibility purposes. As future work we want to explore other more ambitious schemes that take advantage of the speed and simplicity of our approach. One interesting approach would be applications in crowdsourcing platforms where calculations could be performed locally (light weighted edge computing) and then sync sporadically with a master node that would act as a global active learner.

\bibliographystyle{unsrt}  
\bibliography{aaai20.bib}
\end{document}